\documentclass{article}

\usepackage[letterpaper,margin=1in]{geometry}
\usepackage{parskip}
\usepackage[hidelinks]{hyperref}
\usepackage{booktabs}
\usepackage{color}
\usepackage{natbib}
\usepackage{amsmath}
\usepackage{amssymb}
\usepackage{stackengine}
\usepackage{multicol}
\usepackage{tikz-qtree}
\usepackage{tabto}
\usepackage{gb4e} \noautomath

\exewidth{(00)}

\newcommand{\mystack}[2]{{\small$\left\{\stackanchor{\vphantom{ly}#1}{\vphantom{ly}#2}\right\}$}}
\newcommand{\gustack}[2]{\mystack{\phantom{*}#1}{*#2}}
\newcommand{\ugstack}[2]{\mystack{*#1}{\phantom{*}#2}}

\newcommand{\marker}[1]{\fbox{\small\textsc{#1}}}

\title{\citet{KalliniEtAl24} do not compare impossible languages with constituency-based ones}
\author{Tim Hunter\\\texttt{timhunter@ucla.edu}}

\begin{document}
\maketitle

A central goal of linguistic theory, since at least \citet[p.25]{Chomsky65}, has been to find a precise characterization 
of the notion ``possible human language''. Researchers have pursued this goal by attempting to identify a kind of computational 
device that is capable of describing all \emph{and only} the possible human languages, i.e.~those languages that can be 
acquired by a typically developing human child. To the extent that a particular kind of computational device meets this 
goal, it constitutes a plausible hypothesis about the mental machinery that underlies the human capacity for language.

The success of recent ``large language models'' (LLMs) in NLP applications raises the possibility that LLMs might be 
devices that meet this goal. They have been found to be remarkably successful at tasks that, let us grant --- 
controversially, but innocuously for present purposes --- 
require learning certain human languages in a relevant sense. 
The other side of the coin, however, is whether LLMs are similarly successful at learning languages that humans cannot, 
i.e.~``humanly impossible languages''. If they are, this would tell against the hypothesis that human linguistic capacities 
take a form that resembles an LLM.

\citet{KalliniEtAl24} cite a number of claims to the effect that LLMs will successfully learn such impossible languages, 
and set out to test this. They develop a set of synthetic languages with properties that are unlike what has been observed 
in any human language, and find that ``GPT-2 struggles to learn impossible languages when compared to English as a control, 
challenging the core claim''~(p.14691). 
The most interesting impossible languages, and the ones that \citeauthor{KalliniEtAl24}~address most extensively in their paper, 
are those that involve count-based rules. 
Sentences of the language called \textsc{WordHop}, for example, are like sentences of English except that inflectional 
affixes on verbs are replaced with distinguished marker tokens (\marker{s} for singular, \marker{p} for plural) which 
appear to the right of the (uninflected) verb, separated by exactly four words; see \eqref{eg:languages}. 
For a minimal comparison with \textsc{WordHop}, \citeauthor{KalliniEtAl24}~also construct a minor variant of English 
called \textsc{NoHop}, which uses the same distinguished markers but places them immediately adjacent to the 
verb.
\begin{exe}
\ex \label{eg:languages}
    \raisebox{0.8\baselineskip}{
    \begin{tabular}[t]{lll}
    \toprule
                        & Singular agreement example                        & Plural agreement example                              \\
    \midrule
    English             & He cleans his very messy bookshelf .              & They clean his very messy bookshelf .                 \\
    \textsc{WordHop}    & He clean his very messy bookshelf \marker{s} .    & They clean his very messy bookshelf \marker{p} .      \\
    \textsc{NoHop}      & He clean \marker{s} his very messy bookshelf .    & They clean \marker{p} his very messy bookshelf .      \\
    \bottomrule
    \end{tabular}
    }
\end{exe}

It is widely agreed that the count-based placement of the \marker{s} and \marker{p} markers in \textsc{WordHop} is indeed 
outside the bounds of ``possible human languages'' (whereas \textsc{NoHop}, being essentially analogous to English, is not), and 
\citeauthor{KalliniEtAl24}'s results show that GPT-2 is less successful at learning \textsc{WordHop} than \textsc{NoHop}. 
This finding is presented as the main challenge to the claims that GPT-2 models are insufficiently human-like.

The comparison between \textsc{WordHop} and \textsc{NoHop}, however, does not actually test the critical point. 
The problem, to a first approximation, is a confound between whether a rule is count-based and whether that 
rule creates non-adjacent dependencies: the comparison is between \emph{adjacency} and \emph{count-based non-adjacency}. 
The crucial observation that linguists have repeatedly remarked on regarding count-based non-adjacent dependencies is 
their absence relative to \emph{constituency-based non-adjacent} dependencies, not relative to adjacent dependencies. 
The corresponding claim about the human language faculty is that it can naturally accommodate or express constituency-based 
non-adjacent dependencies to a degree that does not hold for count-based non-adjacent dependencies. 
It would be interesting to know whether LLMs show this same asymmetry, but a comparison between 
\textsc{WordHop} and \textsc{NoHop} sheds no light on this question.

In section~\ref{sec:constituency} I will rehearse some standard arguments illustrating the difference between 
count-based and constituency-based rules. 
With some specifics of the relevant phenomena in hand, section~\ref{sec:inflection_system} lays out more carefully 
why the comparison between \textsc{WordHop} and \textsc{NoHop} misses the mark. 
This logic will lead to some suggestions for more appropriate comparisons in section~\ref{sec:better_comparison}.

\section{Review of the underlying issues}
\label{sec:constituency}

The frequently-used example of question-formation in English provides a relevant entry point for 
illustrating the issues.\footnotemark{} 
Consider the relationship that the sentences in \eqref{eg:sai_simple:declarative} and \eqref{eg:sai_pp:declarative} 
stand in to their corresponding yes-no questions. The question form of \eqref{eg:sai_simple:declarative} consists of the same 
words rearranged, as in \eqref{eg:sai_simple:interrog}; we can describe the rearrangement by saying that the word `will' 
has been displaced to the front of the sentence. One could imagine that this was an instance of a count-based rule that 
formed questions by displacing the third word of a sentence, but we can see that this is not the case because applying 
this count-based rule to \eqref{eg:sai_pp:declarative} yields \eqref{eg:sai_pp:move_third}. The actual rule under 
investigation somehow yields \eqref{eg:sai_pp:interrog}, where it is the sixth word that is displaced.

\begin{minipage}[t]{0.4\textwidth}
\begin{exe}
\ex \label{eg:sai_simple}
    \begin{xlist}
    \ex[ ]{\label{eg:sai_simple:declarative}    The dog will bark}
    \ex[ ]{\label{eg:sai_simple:interrog}       Will the dog bark?}
    \end{xlist}
\end{exe}
\end{minipage}
\begin{minipage}[t]{0.5\textwidth}
\begin{exe}
\ex \begin{xlist}
    \ex[ ]{\label{eg:sai_pp:declarative}        The dog in the corner will bark}
    \ex[*]{\label{eg:sai_pp:move_third}         In the dog the corner will bark?}            
    \ex[ ]{\label{eg:sai_pp:interrog}           Will the dog in the corner bark?}
    \end{xlist}
\end{exe}
\end{minipage}
\vskip 0.5\baselineskip

\footnotetext{This argument has appeared in numerous places, virtually unchanged, going back to at least \citet[pp.26--29]{Chomsky71}. 
\citet{Freidin91} gives a version that particularly emphasizes the contrast with count-based rules. 
Other sources include \citet[pp.30--33]{Chomsky75-ROL}, \citet[pp.39--40]{Chomsky80-Piaget} and \citet[pp.41--45]{Chomsky88}. 
For textbook expositions, see e.g.~\citet[pp.156--168]{AkmajianEtAl01}, \citet[pp.5--7]{Lasnik00} and \citet[pp.31--34]{Radford88}. 
Many of these sources discuss this question-formation rule as part of a ``poverty of stimulus'' argument, which need not concern 
us here: what's relevant here is just the initial point that \emph{linguists} can test and disprove hypothesized count-based rules, 
not the subsequent question of how or why language-learners converge on the non-count-based rules that they do.}

Considering now \eqref{eg:sai_complex:declarative}, the question-forming rule displaces neither the third word nor 
the sixth word (which would yield \eqref{eg:sai_complex:third_word} and \eqref{eg:sai_complex:sixth_word} respectively). 
What \eqref{eg:sai_simple:interrog} and \eqref{eg:sai_pp:interrog} have in common is that in both cases the displaced 
word is `will', and this also holds of the desired form \eqref{eg:sai_complex:interrog} --- where the displaced `will' was the 
eighth word. But the rule under investigation somehow excludes moving the other `will', the fourth word of 
\eqref{eg:sai_complex:declarative}, to produce \eqref{eg:sai_complex:first_is}.
\begin{exe}
\ex \label{eg:sai_complex}
    \begin{xlist}
    \ex[ ]{\label{eg:sai_complex:declarative} The dog that will chase the cat will bark}
    \ex[*]{\label{eg:sai_complex:third_word}  That the dog will chase the cat will bark?}     
    \ex[*]{\label{eg:sai_complex:sixth_word}  The the dog that will chase cat will bark?}     
    \ex[ ]{\label{eg:sai_complex:interrog}    Will the dog that will chase the cat bark?}     
    \ex[*]{\label{eg:sai_complex:first_is}    Will the dog that chase the cat will bark?}     
    \end{xlist}
\end{exe}

And it is not as simple as always moving the last/rightmost occurrence of `will' (or more generally, an auxiliary verb), 
as illustrated by the pattern in \eqref{eg:sai_complex_other}.
\begin{exe}
\ex \label{eg:sai_complex_other}
    \begin{xlist}
    \ex[ ]{The dog in the corner will chase the dog that will bark}
    \ex[ ]{Will the dog in the corner chase the dog that will bark?}
    \ex[*]{Will the dog in the corner will chase the dog that bark?}
    \end{xlist}
\end{exe}

The operative rule cannot be formulated in count-based terms, i.e.~no description of the form ``the $n$th word 
of the sentence'' or ``the $n$th occurrence of `will' from the end of the sentence'' will consistently pick out the 
word that is to be displaced.

The correct generalization \emph{can} be expressed in terms of hierarchical constituency: given the structural analyses in 
\eqref{eg:pre_sai_trees} for the declaratives in \eqref{eg:sai_complex} and \eqref{eg:sai_complex_other}, the displaced 
word is the Aux that is the granddaughter of the root S node in the template in \eqref{eg:sai_template}.

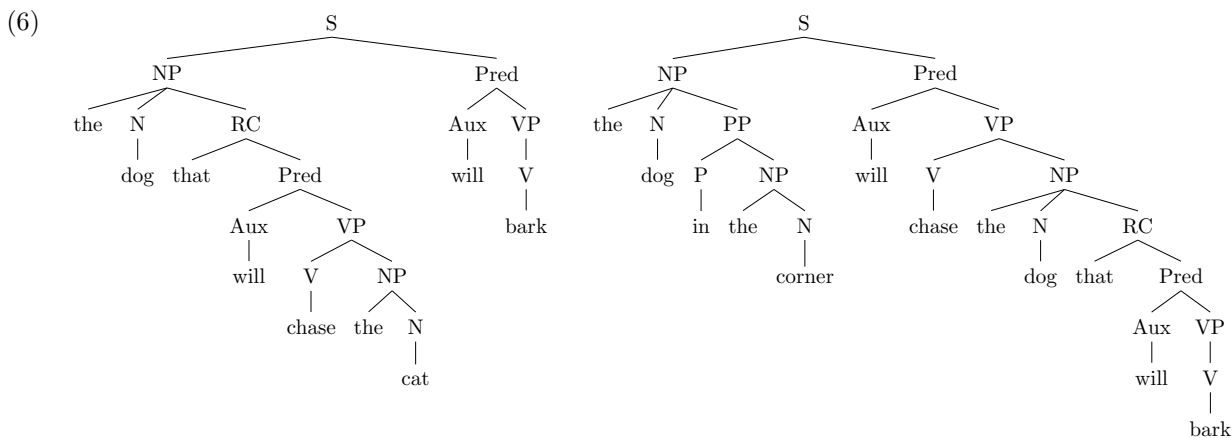
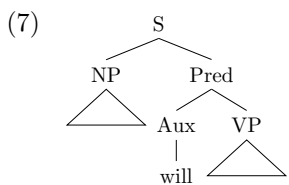
\begin{exe}
\ex \label{eg:pre_sai_trees}
    \tikz[baseline,scale=0.8, level distance=2\baselineskip]{\Tree [.S [.NP the [.N dog ] [.RC that [.Pred [.Aux will ] [.VP [.V chase ] [.NP the [.N cat ] ] ] ] ] ] [.Pred [.Aux will ] [.VP [.V bark ] ] ] ]}
    \hskip 1em
    \tikz[baseline,scale=0.8, level distance=2\baselineskip]{\Tree [.S [.NP the [.N dog ] [.PP [.P in ] [.NP the [.N corner ] ] ] ] [.Pred [.Aux will ] [.VP [.V chase ] [.NP the [.N dog ] [.RC that [.Pred [.Aux will ] [.VP [.V bark ] ] ] ] ] ] ] ]}
\end{exe}

\begin{exe}
\ex \label{eg:sai_template}
    \tikz[baseline, scale=0.8, level distance=2\baselineskip]{\Tree [.S [.NP \edge[roof]; {\hspace*{3em}} ] [.Pred [.Aux will ] [.VP \edge[roof]; {\hspace*{3em}} ] ] ]}
\end{exe}

This example from English is entirely representative: patterns like this that conform to a constituency-based rule, 
but where no count-based characterization has been found, are ubiquitous in natural languages. 
And the reverse situation, where a pattern follows a count-based rule but has no constituency-based characterization, 
is unheard of. 
The conventional linguistic explanation for this striking asymmetry is that (languages with) count-based rules are ``humanly 
impossible'' --- outside the capacity of the mental faculties that are recruited in naturalistic language development.\footnotemark{} 
Of course, given a simple enough artificial grammar-learning experiment, a human may well show some success at learning and applying a 
count-based rule, perhaps by recruiting \emph{other} mental faculties to the task; somewhat similarly, a proponent of the idea that 
LLMs embody a human-like ill-suitedness to count-based rules is not committed to the prediction that an LLM will always show 
precisely zero evidence of having extracted any count-based rule from training data. 
Rather than any raw measure of successful learning of any single kind of rule, the critical issue is an asymmetry between 
count-based and constituency-based rules.

\footnotetext{The idea is not that a count-based language would ``die out'' because of a failure on the part of 
human learners to perpetuate it; rather, the idea is that no human's linguistic development would ever give rise 
to such a language in the first place.}

Testing for such an asymmetry obviously requires controlling for other factors. 
While the rule for the placement of the \marker{s} and \marker{p} markers in \citeauthor{KalliniEtAl24}'s \textsc{WordHop} 
is a canonical example of a count-based rule --- the kind that turns out to be insufficient to describe the 
pattern in \eqref{eg:sai_simple}--\eqref{eg:sai_complex_other} --- the rule for placing these markers in \textsc{NoHop} 
is not an appropriately representative constituency-based rule to compare it against. 
The \textsc{NoHop} rule is extremely simple: the marker should be placed immediately after the verb. 
It's true that the full-fledged English system of verbal inflections involves crucially constituency-based rules, 
which are in fact closely intertwined with the classic phenomena in \eqref{eg:sai_simple}--\eqref{eg:sai_complex_other} above, 
and one of the configurations that this system produces is the one illustrated in \eqref{eg:languages}, with the 
inflected verb `clean\underline{s}'. But the constituency-based parts of that system are not probed by a 
comparison between \textsc{WordHop} and \textsc{NoHop}, which differ \emph{only} in whether the \marker{s} and 
\marker{p} markers are separated from the verb by four words or zero words.

To flesh out this point, section~\ref{sec:inflection_system} illustrates some of the constituency-based rules governing 
English verbal inflections that turn out to be independent of the differences between \textsc{WordHop} and \textsc{NoHop}. 
This will then lead to a proposal for a more appropriate comparison in section~\ref{sec:better_comparison}.

\section{Constituency and English verbal inflections}
\label{sec:inflection_system}

A mistaken impression that \textsc{NoHop} can serve as a representative of constituency-based rules might arise, 
in part, from the fact that the behaviour of verbal inflections is intertwined with the question-forming rule that 
is used in the classical illustration of constituency-sensitivity rehearsed in section~\ref{sec:constituency}. 

This connection can be established by observing that these inflections (e.g.~the suffixes in `clean\underline{s}' 
and `clean\underline{ed}') do not co-occur with words like `will' that are displaced by the question-forming rule. 
A finite clause must include either one of these inflections or a word that behaves like `will' 
(e.g.~`may', `must', `can'), but not both.
\begin{multicols}{3}
\begin{exe}
\ex \label{eg:aux_aff_none}
    \begin{xlist}
    \ex[*]{He clean}
    \ex[ ]{He \underline{will} clean    \label{eg:aux_aff_none:will}}
    \ex[ ]{He \underline{may} clean     \label{eg:aux_aff_none:may}}
    \end{xlist}
\ex \label{eg:aux_aff_s}
    \begin{xlist}
    \ex[ ]{He clean\underline{s}        \label{eg:aux_aff_s:none}}
    \ex[*]{He \underline{will} clean\underline{s}}
    \ex[*]{He \underline{may} clean\underline{s}}
    \end{xlist}
\ex \label{eg:aux_aff_ed}
    \begin{xlist}
    \ex[ ]{He clean\underline{ed}       \label{eg:aux_aff_ed:none}}
    \ex[*]{He \underline{will} clean\underline{ed}}
    \ex[*]{He \underline{may} clean\underline{ed}}
    \end{xlist}
\end{exe}
\end{multicols}

So we have effectively identified a three-way dependency between (i)~the sentence-initial position occupied by `will' 
in the questions in section~\ref{sec:constituency}, (ii)~the position occupied by `will' in 
non-questions, in section~\ref{sec:constituency} and in 
\eqref{eg:aux_aff_none}--\eqref{eg:aux_aff_ed}, and (iii)~the position occupied by the 
inflectional affixes in \eqref{eg:aux_aff_none}--\eqref{eg:aux_aff_ed}. 
This can be formalized in various ways (see \citet{Chomsky57} for the original analysis along these lines\footnotemark{}), 
but the dotted lines in \eqref{eg:three_way_dependency} convey the key idea in a way that will suffice for our purposes here. 
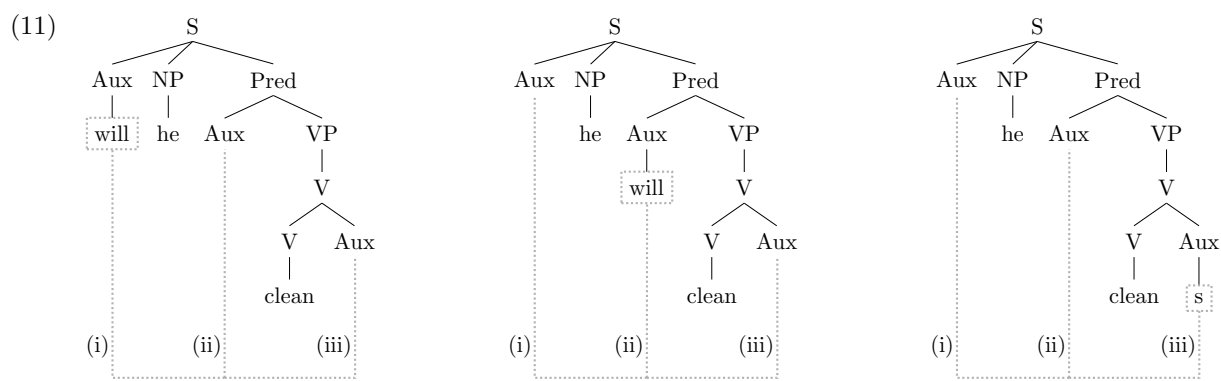
\begin{exe}
\ex \label{eg:three_way_dependency}
    \begin{tikzpicture}[baseline, scale=0.85, transform shape, level distance=2\baselineskip]
    \begin{scope}
        \Tree [.\node(root){S}; [.Aux \node(aux1){will}; ] [.NP he ] [.Pred \node(aux2){Aux}; [.VP [.V [.V clean ] \node(aux3){Aux}; ] ] ] ]
        \coordinate[yshift=-30pt] (mid) at (aux2 |- \subtreeof{root}.south) ;
        \draw[thick, densely dotted, gray!60] (mid) -| (aux1) ;
        \draw[thick, densely dotted, gray!60] (mid) -- (aux2) ;
        \draw[thick, densely dotted, gray!60] (mid) -| (aux3) ;
        \draw[thick, densely dotted, gray!60] (aux1.north west) rectangle (aux1.south east) ;
        \node[anchor=north east, inner sep=1pt, yshift=-10pt] at (\subtreeof{root}.south -| aux1) {(i)};
        \node[anchor=north east, inner sep=1pt, yshift=-10pt] at (\subtreeof{root}.south -| aux2) {(ii)};
        \node[anchor=north east, inner sep=1pt, yshift=-10pt] at (\subtreeof{root}.south -| aux3) {(iii)};
    \end{scope}
    \begin{scope}[xshift=0.4\textwidth]
        \Tree [.S \node(aux1){Aux}; [.NP he ] [.Pred [.Aux \node(aux2){will}; ] [.VP [.V [.V clean ] \node(aux3){Aux}; ] ] ] ]
        \coordinate[yshift=-30pt] (mid) at (aux2 |- \subtreeof{root}.south) ;
        \draw[thick, densely dotted, gray!60] (mid) -| (aux1) ;
        \draw[thick, densely dotted, gray!60] (mid) -- (aux2) ;
        \draw[thick, densely dotted, gray!60] (mid) -| (aux3) ;
        \draw[thick, densely dotted, gray!60] (aux2.north west) rectangle (aux2.south east) ;
        \node[anchor=north east, inner sep=1pt, yshift=-10pt] at (\subtreeof{root}.south -| aux1) {(i)};
        \node[anchor=north east, inner sep=1pt, yshift=-10pt] at (\subtreeof{root}.south -| aux2) {(ii)};
        \node[anchor=north east, inner sep=1pt, yshift=-10pt] at (\subtreeof{root}.south -| aux3) {(iii)};
    \end{scope}
    \begin{scope}[xshift=0.8\textwidth]
        \Tree [.S \node(aux1){Aux}; [.NP he ] [.Pred \node(aux2){Aux}; [.VP [.V [.V clean ] [.Aux \node(aux3){s}; ] ] ] ] ]
        \coordinate[yshift=-30pt] (mid) at (aux2 |- \subtreeof{root}.south) ;
        \draw[thick, densely dotted, gray!60] (mid) -| (aux1) ;
        \draw[thick, densely dotted, gray!60] (mid) -- (aux2) ;
        \draw[thick, densely dotted, gray!60] (mid) -| (aux3) ;
        \draw[thick, densely dotted, gray!60] (aux3.north west) rectangle (aux3.south east) ;
        \node[anchor=north east, inner sep=1pt, yshift=-10pt] at (\subtreeof{root}.south -| aux1) {(i)};
        \node[anchor=north east, inner sep=1pt, yshift=-10pt] at (\subtreeof{root}.south -| aux2) {(ii)};
        \node[anchor=north east, inner sep=1pt, yshift=-10pt] at (\subtreeof{root}.south -| aux3) {(iii)};
    \end{scope}
    \end{tikzpicture}
\end{exe}

\footnotetext{For textbook expositions see e.g.~\citet[pp.259--300]{FromkinEtAl00}, \citet[pp.246--271]{Carnie07}, 
\citet[pp.66--86]{Lasnik00}, \citet[pp.144-164]{Freidin92}.}

To complete the picture, notice that the question-forming rule does not make any distinction between the affixes that 
appear in position~(iii) in declaratives and the words like `will' that appear in position~(ii): the affixes are 
also displaced to position~(i) in questions, where their pronunciation is supported by a form of the dummy verb `do'.
\begin{multicols}{2}
\begin{exe}
\ex \begin{xlist}
    \ex Does he clean?  \tabto{0.6\linewidth} (cf.~\eqref{eg:aux_aff_s:none})
    \ex Did he clean?   \tabto{0.6\linewidth} (cf.~\eqref{eg:aux_aff_ed:none})
    \ex Will he clean?  \tabto{0.6\linewidth} (cf.~\eqref{eg:aux_aff_none:will})
    \ex May he clean?   \tabto{0.6\linewidth} (cf.~\eqref{eg:aux_aff_none:may})
    \end{xlist}
\end{exe}
\end{multicols}

No matter how these details are formalized, the crucial and uncontroversial point is that these three interdependent 
positions are identified in constituent-based terms, not count-based terms. 
We saw in section~\ref{sec:constituency} that the relationship between positions~(i) and~(ii) 
is not defined via a number of intervening words, but rather with reference to the hierarchical structure. 
Similarly, although the word that an affix in position~(iii) attaches to has been adjacent to position~(ii) in all the 
examples so far, this is not true in general: additional words can intervene here too, as illustrated by 
\eqref{eg:affix_hopping_intervenors}. The presence of a direct object after the verb in these sentences also demonstrates 
that position~(iii) can not be defined linearly as ``the end of the string''. 
\begin{exe}
\ex \label{eg:affix_hopping_intervenors}
    \begin{xlist}
    \ex He \underline{will} without doubt clean his very messy bookshelf.
    \ex He without doubt clean\underline{s} his very messy bookshelf.
    \end{xlist}
\end{exe}

Furthermore, although the discussion in section~\ref{sec:constituency} emphasized only the hierarchical 
determination of the auxiliary that should be displaced to the front of the sentence, this target position is in fact defined 
in hierarchical terms too: \eqref{eg:sai_non_initial} shows examples of questions where `will' is in position~(i) despite 
not being sentence-initial.\footnote{Relevant examples here are restricted by the fact that, in most varieties of English, 
subject-auxiliary inversion only occurs in matrix clauses. In some varieties spoken in Ireland, for example, the same operation 
applies in embedded clauses, yielding examples like `I wonder will he clean it?' \citep{McCloskey92,McCloskey06,Henry95}.}
\begin{exe}
\ex \label{eg:sai_non_initial}
    \begin{xlist}
    \ex Which very messy bookshelf \underline{will} he clean?
    \ex How \underline{will} he clean his very messy bookshelf?
    \ex Though his bookshelf is very messy, \underline{will} he clean it?
    \end{xlist}
\end{exe}

Another way in which English verbal inflections are intertwined with crucially hierarchical notions concerns number 
agreement with the subject; recall the two columns in \eqref{eg:languages}. 
There is a single hierarchically-defined position that the agreement-controlling noun `gift(s)' occupies in all of the 
examples in \eqref{eg:agreement_subject_pp}--\eqref{eg:agreement_subject_poss}. 
The rule needs to pick out the second word (and the first of the two nouns) in \eqref{eg:agreement_subject_pp}, but 
the third word (and the second of the two nouns) in \eqref{eg:agreement_subject_poss}, so again no count-based formulation is possible.
\begin{multicols}{2}
\begin{exe}
\ex \label{eg:agreement_subject_pp}
    \begin{xlist}
    \ex The gift from the alumnus \gustack{matters}{matter}
    \ex The gift from the alumni \gustack{matters}{matter}
    \ex The gifts from the alumnus \ugstack{matters}{matter}
    \ex The gifts from the alumni \ugstack{matters}{matter}
    \end{xlist}
\ex \label{eg:agreement_subject_poss}
    \begin{xlist}
    \ex The alumnus's gift \gustack{matters}{matter}
    \ex The alumni's gift \gustack{matters}{matter}
    \ex The alumnus's gifts \ugstack{matters}{matter}
    \ex The alumni's gifts \ugstack{matters}{matter}
    \end{xlist}
\end{exe}
\end{multicols}

Both of these phenomena have (with good reason) been prominent test cases in work investigating connectionist systems' 
treatment of constituency-based generalizations. 
Studies using the question-forming rule as a probe into this issue include \citet{FrankMathis07}, \citet{McCoyEtAl20} and \citet{WarstadtBowman20}, 
and those using subject-verb agreement include \citet{LinzenEtAl16}, \citet{KuncoroEtAl18} and \citet{LakretzEtAl21}.
And as illustrated in this section, the crucially constituency-sensitive rules underlying both of these phenomena 
bear on the distribution of English inflected verb forms (e.g.~`clean\underline{s}' and `clean\underline{ed}') that 
\citeauthor{KalliniEtAl24} manipulate in order to create \textsc{WordHop} and \textsc{NoHop}. 
But English sentences with those inflected verb forms are a shared ``starting point'' for these two artificial languages, 
which differ only in whether the \marker{s} and \marker{p} markers occur in the hierarchically-defined position~(iii) 
or at a count-based offset from \emph{that} position. 
The constituency-based patterns in which verbal inflections participate --- the dependency between the three 
positions illustrated in \eqref{eg:three_way_dependency}, and the hierarchical determination of the controller of 
agreement illustrated in \eqref{eg:agreement_subject_pp}--\eqref{eg:agreement_subject_poss} --- are 
irrelevant for any \emph{comparison} between \textsc{WordHop} and \textsc{NoHop}. 
\textsc{WordHop} contains just as much constituency-based question-formation, and just as much hierarchically-sensitive 
agreement, as \textsc{NoHop} does. 
A comparison between the two just amounts to a comparison between the count-based displacement in \textsc{WordHop}, 
and the absence of any analogous displacement in \textsc{NoHop}.

\section{Towards a better comparison: counting vs.~constituency}
\label{sec:better_comparison}

The problem with the comparison between \textsc{WordHop} and \textsc{NoHop} is that the count-based rule in \textsc{WordHop} 
is not the counterpart of any constituency-based rule in \textsc{NoHop}. There are two ways we might seek to rectify this. 
The first is to keep \textsc{WordHop} as our representative count-based language, and introduce a constituency-based rule 
to be the necessary counterpart: compare \textsc{WordHop}, where the \marker{s} and \marker{p} markers are placed at a 
count-based offset from position~(iii), against a new synthetic language where these markers are placed at a 
constituency-based offset from position~(iii). 
The second possibility is to keep \textsc{NoHop} as our representative constituency-based language, and \emph{replace} 
one of the constituency-based rules governing the placement of the \marker{s} and \marker{p} markers with a count-based rule. 
Either route leads to some subtle issues that remain to be worked out, and my aim here is only to advance the discussion in a 
way that lays out the logic and clarifies what is needed, not to fully resolve the isssues that arise.

To introduce a constituency-based counterpart to \textsc{WordHop}'s count-based rule, we need to identify a 
hierarchically-defined offset from position~(iii) where markers would be placed in the new artificial language. 
Suppose we choose the right edge of the sister constituent of position~(iii)'s parent V node; this will be the right edge 
of the direct object, in many cases. (There is no such constituent in the minimal diagrams in \eqref{eg:three_way_dependency}, 
but notice the relevant NP constituents in \eqref{eg:pre_sai_trees}.) 
Then we would have a comparison between the count-based language illustrated in \eqref{eg:comparison_count1} (unchanged from 
\citeauthor{KalliniEtAl24}'s \textsc{WordHop}) and the constituency-based language illustrated in \eqref{eg:comparison_constituency1}.
\begin{multicols}{2}
\begin{exe}
\ex \label{eg:comparison_count1}
    Count-based (unchanged from \textsc{WordHop})
    \begin{xlist}
    \ex He clean his very messy bookshelf \marker{s}
    \ex He clean the bookshelf with glee \marker{s}
    \ex He clean it with a big \marker{s} red broom
    \ex He clean the bookshelf that is \marker{s} messy
    \end{xlist}
\ex \label{eg:comparison_constituency1}
    Constituency-based
    \begin{xlist}
    \ex He clean his very messy bookshelf \marker{s}
    \ex He clean the bookshelf \marker{s} with glee
    \ex He clean it \marker{s} with a big red broom
    \ex He clean the bookshelf that is messy \marker{s}
    \end{xlist}
\end{exe}
\end{multicols}

One challenge here is that synthesizing examples like those in \eqref{eg:comparison_constituency1} from their English equivalents 
requires settling on an analysis of what counts as a sister of the relevant V node, which will be controversial in some cases; 
and in practical terms, even to the extent that analyses of individual cases are uncontroversial, 
they would likely to difficult to automate.\footnotemark{}

\footnotetext{Under any reasonable assumptions there will be many English examples where no such sister constituent exists, 
and these would need to be excluded from the corpora to keep the comparison balanced --- just as \citeauthor{KalliniEtAl24} 
excluded sentences where an inflected verb was too close to the right edge for their \textsc{WordHop} rule to apply.}

A more subtle concern is whether the pattern in \eqref{eg:comparison_constituency1} is necessarily describable only in terms of 
a constituency-based offset from position~(iii), or whether it has an alternative characterization in terms of a constituency-based 
offset from position~(ii). If the position of the markers in \eqref{eg:comparison_constituency1} can be understood as part of a 
hierarchically-defined dependency with position~(ii), then the pattern in \eqref{eg:comparison_constituency1} would be no 
better than \textsc{NoHop}: the comparison between \eqref{eg:comparison_count1} and \eqref{eg:comparison_constituency1} would 
again be a comparison between the composition of a constituency-based and a count-based offset from position~(ii), and an 
only constituency-based offset from position~(ii). The underlying question here is whether the composition of two 
constituency-based relations is always another valid constituency-based relation. The answer will depend on the details 
of one's theory of linguistically possible dependencies, which remains an active research topic. 
(It may bear repeating here that the exclusion of count-based dependencies is not one of the points of disagreement.)

Consider now the other route, where we pit a count-based rule against one of the existing constituency-based rules 
underlying \textsc{NoHop}. Let's suppose the relevant count-based rule placed the \marker{s} and \marker{p} markers 
at a four-word offset from position~(ii), as a counterpart to the standard hierarchically-defined relationship between 
position~(ii) and position~(iii) illustrated in \eqref{eg:three_way_dependency}. 
This comparison is illustrated in \eqref{eg:comparison_count2} and \eqref{eg:comparison_constituency2}.
\begin{multicols}{2}
\begin{exe}
\ex \label{eg:comparison_count2}
    Count-based
    \begin{xlist}
    \ex He clean his messy bookshelf \marker{s}
    \ex He always clean his messy \marker{s} bookshelf
    \ex He without doubt clean it \marker{s}
    \ex He clean it with a \marker{s} broom
    \end{xlist}
\ex \label{eg:comparison_constituency2}
    Constituency-based (unchanged from \textsc{NoHop})
    \begin{xlist}
    \ex He clean \marker{s} his messy bookshelf
    \ex He always clean \marker{s} his messy bookshelf
    \ex He without doubt clean \marker{s} it
    \ex He clean \marker{s} it with a broom
    \end{xlist}
\end{exe}
\end{multicols}

The necessary syntactic analysis for synthesizing examples like \eqref{eg:comparison_count2} only requires identifying 
position~(ii) (the position of auxiliary verbs in standard declaratives), which is likely less controversial than the 
issues that arise for \eqref{eg:comparison_constituency1} surrounding sister constituents of the verb.

A questionable aspect of the comparison between \eqref{eg:comparison_count2} and \eqref{eg:comparison_constituency2} is 
that, in the constituency-based pattern, the word immediately preceding the marker is always of the same category 
(namely a verb), whereas in the count-based pattern the words preceding the marker are heterogeneous in 
syntactic category. (This is also a characteristic of the comparison between \textsc{WordHop} and \textsc{NoHop}.) 
This could be thought to make the constituency-based pattern more ``predictable'' or ``simple'' than the count-based 
one in a sense that we would like to control for. 
Notice that this consistency of an adjacent category is not a general property of constituency-based rules: 
in \eqref{eg:comparison_constituency1} we see the marker follow `bookshelf', `it' and `messy', which belong 
to distinct syntactic categories. 
Rather it is a consequence of the fact that the rule that relates position~(ii) and position~(iii) in English 
(``affix hopping'') is somewhat anomalous in ways that lead to divided opinions over whether it is best 
considered a morphological or syntactic rule \citetext{e.g.~\citealp[pp.134--138]{HalleMarantz93}; \citealp[pp.584--591]{EmbickNoyer01}}.

As mentioned above, I make no attempt to resolve all these issues here; the main goal of presenting 
\eqref{eg:comparison_count1} vs.~\eqref{eg:comparison_constituency1} and 
\eqref{eg:comparison_count2} vs.~\eqref{eg:comparison_constituency2} is to 
lay out the logic of what would make an informative comparison between count-based and constituency-based 
rules, and in doing so clarify the earlier critiques of the comparison that \citeauthor{KalliniEtAl24} report.

\section{Conclusion}

In natural languages, words that are linked by some grammatical dependency do not always appear adjacent to each other. 
What linguists have taken to be striking is that the rules governing these non-adjacent configurations of co-dependent 
words are never describable in terms of (relative) numerical positions in the string; instead, the positions involved 
are characterized in constituency-based terms. 
This is hypothesized to be a consequence of an important difference in the status of count-based versus constituency-based 
rules in the human mind. \citeauthor{KalliniEtAl24} present their comparison between \textsc{WordHop} and \textsc{NoHop} 
as a test of whether GPT-2 shows an analogous asymmetry, but these two artificial languages do not differ in the 
appropriate way for this interpretation: the count-based rule in \textsc{WordHop} has no counterpart (constituency-based 
or otherwise) in \textsc{NoHop}, and so differences in learning success reflect the presence of this additional rule, not 
an asymmetry between two kinds of rules.

Of course, nothing I have said amounts to any claim about the underlying question of whether an LLM might 
exhibit a human-like asymmetry between count-based and constituency-based rules. The claim here is just that the 
experiments reported by \citeauthor{KalliniEtAl24} leave the issue untouched.

\bibliographystyle{apalike}
\bibliography{linguistics}

\end{document}